\pdfoutput=1
\documentclass[11pt]{article}
\usepackage{EMNLP2022}

\usepackage{times}
\usepackage{latexsym}
\usepackage[T1]{fontenc}
\usepackage[utf8]{inputenc}
\usepackage{microtype}
\usepackage{inconsolata}

\usepackage{booktabs, tabularx}
\usepackage{graphics}
\usepackage{amsmath}
\usepackage{soul,array,multirow,graphicx}
\usepackage{caption}
\usepackage{subcaption}
\usepackage{tikz-dependency}
\usepackage{pgfplots}
\pgfplotsset{compat=1.17}
\usepackage{CJKutf8}
\usepackage[ruled,vlined]{algorithm2e}
\usepackage{hyperref}
\usepackage{amssymb}

\newcommand{\flag}[1]{#1}
\newcommand{\modelname}{CubeRE}
\newcommand{\dataname}{HyperRED}

\newcommand{\declarelogo}[0]{\includegraphics[height=.02\textwidth]{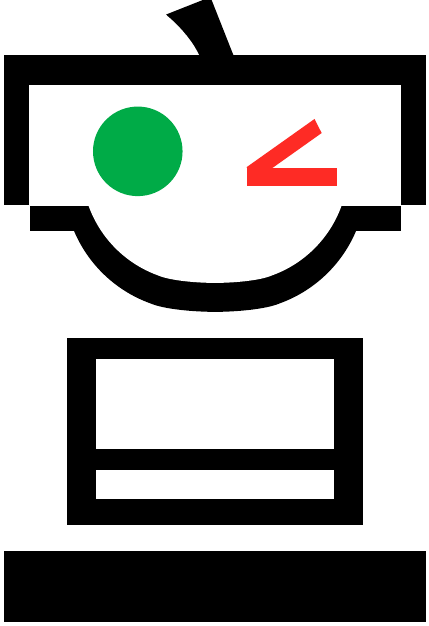}}

\definecolor{commentcolor}{RGB}{110,154,155}   
\newcommand{\PyComment}[1]{\ttfamily\textcolor{commentcolor}{\# #1}}  
\newcommand{\PyCode}[1]{\ttfamily\textcolor{black}{#1}} 

\title{
{A Dataset for Hyper-Relational Extraction and a Cube-Filling Approach}
}

\author{
\textbf{
Yew Ken Chia\thanks{$^{*}$Yew Ken is a student under the Joint PhD Program between Alibaba and SUTD. 
}
~\textsuperscript{\rm 1,${\declarelogo}$}\quad
Lidong Bing\thanks{~~Corresponding author.}~\textsuperscript{\rm 1}\quad
Sharifah Mahani Aljunied\textsuperscript{\rm 1}~~~
}\\
\textbf{
Luo Si\textsuperscript{\rm 1}~~~~~~~~~~
Soujanya Poria\textsuperscript{\rm ${\declarelogo}$}\quad} \\
\textsuperscript{\rm 1}DAMO Academy, Alibaba Group~~
\textsuperscript{\rm ${\declarelogo}$} Singapore University of Technology and Design ~~\\
{\tt\{yewken.chia, l.bing, mahani.aljunied, luo.si\}@alibaba-inc.com} \\~~{\tt\{yewken\_chia, sporia\}@sutd.edu.sg}}

\begin{document}
\maketitle
\begin{abstract}

Relation extraction has the potential for large-scale knowledge graph construction, but current methods do not consider the qualifier attributes for each relation triplet, such as time, quantity or location.
The qualifiers form hyper-relational facts which better capture the rich and complex knowledge graph structure.
For example, the relation triplet (Leonard Parker, Educated At, Harvard University) can be factually enriched by including the qualifier (End Time, 1967).
Hence, we propose the task of hyper-relational extraction to extract {more specific and complete facts from text}.
To support the task, we construct HyperRED, a large-scale and general-purpose dataset.
Existing models cannot perform hyper-relational extraction as it requires a model to consider the interaction between three entities.
Hence, we propose \modelname{}, a cube-filling model inspired by table-filling approaches 
and explicitly considers the interaction between relation triplets and qualifiers.
To improve model scalability and reduce negative class imbalance, we further propose a cube-pruning method.
Our experiments show that \modelname{} outperforms strong baselines and reveal possible directions for future research.
Our code and data are available at 
\href{https://github.com/declare-lab/HyperRED}{github.com/declare-lab/HyperRED}.
\end{abstract}

\section{Introduction}

Knowledge acquisition is an open challenge in artificial intelligence research \citep{10.1145/219717.219745}.
The standard form of representing the acquired knowledge is a knowledge graph \citep{HOVY20132}, which has broad applications such as question answering \citep{yih-ma-2016-question, chia-etal-2020-red} and search engines \citep{10.1145/3038912.3052558}.
Relation extraction (RE) is a task that has the potential for large-scale and automated knowledge graph construction by extracting facts from natural language text.
Most relation extraction methods focus on binary relations \citep{bach2007review} which consider the relationship between two entities, forming a relation triplet consisting of the head entity, relation and tail entity respectively.

\begin{figure}[!t]
\centering
\includegraphics[width=1.0\columnwidth]{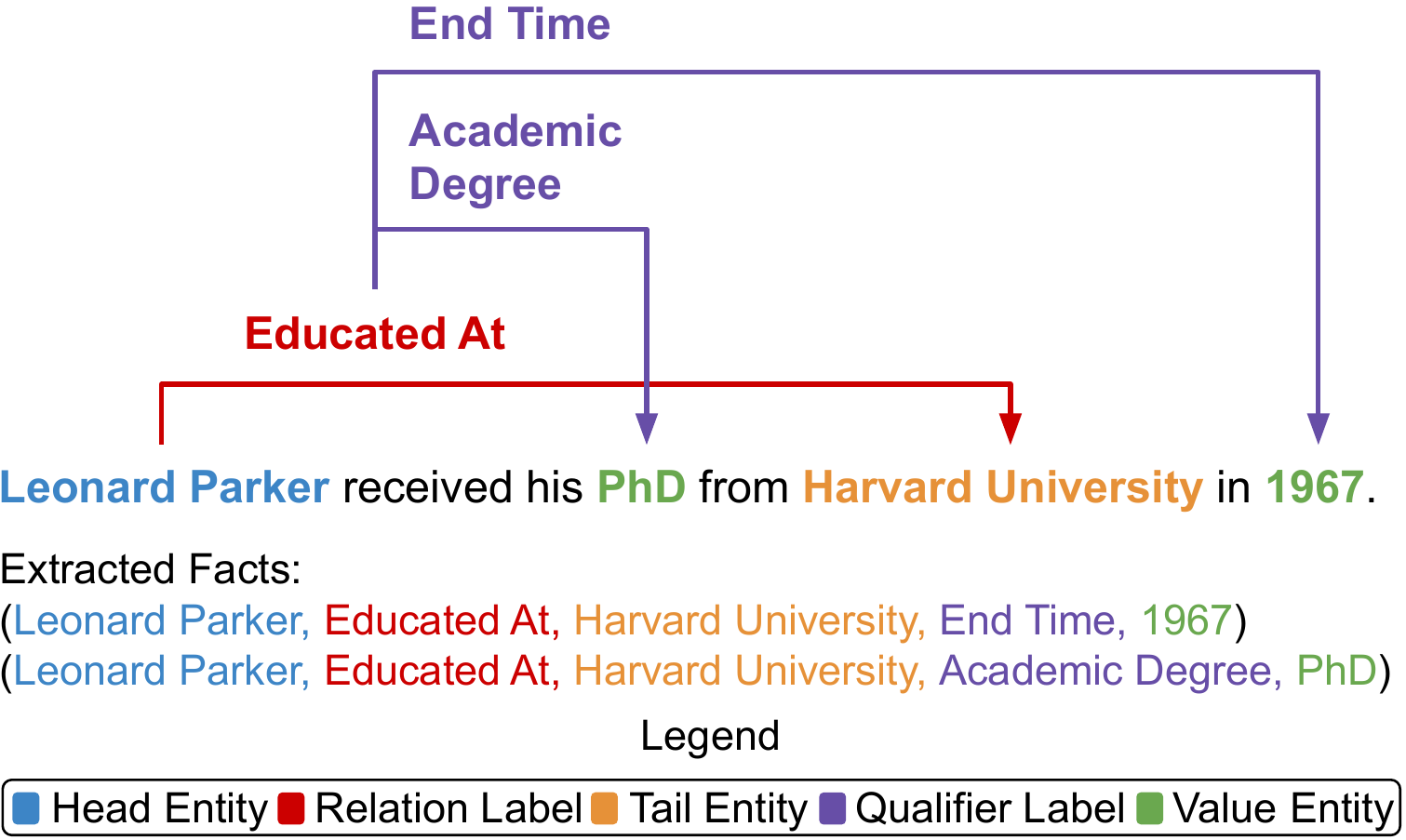}
\caption{
A sample from our \dataname{} dataset for the proposed task of hyper-relational extraction. 
\vspace{-5mm}}
\label{fig:example}
\end{figure}

However, knowledge graphs commonly contain hyper-relational facts \citep{10.1145/3308558.3313414} which have qualifier attributes for each relational triplet, such as time, quantity, or location.
For instance, \citet{10.5555/3060621.3060802} found that the Freebase knowledge graph contains hyper-relational facts for 30\% of entities.
Hence, extracting relation triplets may be an oversimplification of the rich and complex knowledge graph structure.
As shown in Figure \ref{fig:example}, a relation triplet can be attributed to one or more qualifiers, where a qualifier is composed of a qualifier label and value entity.
For example, the relation triplet (\texttt{Leonard Parker}, \textit{Educated At}, \texttt{Harvard University}) can be factually enriched by specifying the 
qualifier of (\textit{End Time}, \texttt{1967}), forming the hyper-relational fact (\texttt{Leonard Parker}, \textit{Educated At}, \texttt{Harvard University}, \textit{End Time}, \texttt{1967}).

Hyper-relational facts generally cannot be simplified into the relation triplet format as the qualifiers are attributed to the triplet as a whole and not targeted at a specific entity in the triplet.
Furthermore, attempting to decompose the hyper-relational structure to an n-ary format would lose the original triplet information and be incompatible with the knowledge graph schema \citep{10.1145/3366423.3380257}.
On the other hand, hyper-relational facts have practical benefits such as improved fact verification \citep{thorne-etal-2018-fever}
and representation learning for knowledge graphs \citep{galkin-etal-2020-message}.
Thus, it is necessary to extract relation triplets together with qualifiers to form hyper-relational facts.

In this work, we propose the task of hyper-relational extraction to jointly extract relation triplets with qualifiers from natural language sentences.
To support the task, we contribute a general-purpose and large-scale hyper-relational extraction dataset (\dataname{}) which is constructed through distant supervision \citep{mintz-etal-2009-distant} and partially refined through human annotation.
Our dataset differs from previous datasets in two distinct ways:
(1) Compared to existing datasets for binary relation extraction \citep{zhang-etal-2017-position, han-etal-2018-fewrel}, \dataname{} enables richer information extraction as it contains qualifiers for each relation triplet in the sentence. 
(2) While datasets for n-ary relation extraction \citep{jia-etal-2019-document} are restricted to the biomedical domain, \dataname{} covers multiple domains and has a hyper-relational fact structure that is compatible with the knowledge graph schema. 

Unfortunately, to the best of our knowledge, there are no existing models for hyper-relational extraction.
Currently, a popular end-to-end method for binary relation extraction is to cast it as a table-filling problem \citep{miwa-sasaki-2014-modeling}.
Generally, a two-dimensional table is used to represent the interaction between any two individual words in a sentence.
However, hyper-relational extraction requires the model to consider the interactions between two entities in the relation triplet, as well as the value entity for the qualifier.
Thus, we extend the table-filling approach to a third dimension, casting it as a cube-filling problem. 
On the other hand, a naive cube-filling approach faces two issues: 
(1) Computing the full cube representation is computationally expensive and does not scale well to longer sequence lengths.
(2) The full cube will be sparsely labeled with a vast majority of entries as negative samples, causing the model to be biased in learning \citep{li-etal-2020-dice} and hence underperform.

To tackle these two issues, we propose a simple yet effective cube-pruning technique that filters the cube entries based on words that are more likely to constitute valid entities.
Our experiments show that cube-pruning significantly improves the computational efficiency and simultaneously improves the extraction performance by reducing the negative class imbalance.
In addition to our cube-filling model which we refer to as \modelname, we also introduce two strong baseline models which include a two-stage pipeline and a generative sequence-to-sequence \citep{NIPS2014_a14ac55a} model. 

In summary, our main contributions include: (1) We propose the task of hyper-relational extraction to extract richer and more complete facts by jointly extracting each relation triplet with the corresponding qualifiers; (2) To support the task, we provide a large-scale and general-purpose dataset known as \dataname.
(3) As there is no existing model for hyper-relational extraction, we propose a cube-filling model known as \modelname, which consistently outperforms baseline extraction methods.

\section{HyperRED: A Hyper-Relational Extraction Dataset}
Our goal is to construct a large-scale and general-purpose dataset for extracting hyper-relational facts from natural language text.
However, it is seldom practical to assume to have an ample amount of high-quality labeled samples in real applications, especially for complex tasks such as information extraction.
Hence, we propose a weakly supervised \citep{10.5555/645634.663209} data setting which enables us to collect a larger and more diverse training set than would be otherwise possible.
To minimize the effect of noisy samples in evaluation, we then perform human annotation for a portion of the collected data and allocate it as the held-out set.
In the following sections, we first introduce the process of collecting the distantly supervised data, followed by the human-annotated data portion. 

\begin{table*}[!t]
    \centering
    \resizebox{1\textwidth}{!}{
    \begin{tabular}{llll}
    \toprule
    \textbf{Type} & \textbf{Proportion} & \textbf{Example Sentence} & \textbf{Hyper-Relational Facts} \\
    \midrule
    Time & 48\% & Tennyson was an ASCAP member from 1950. & (Tennyson, member of, ASCAP, start time, 1950) \\
    \midrule
    Quantity & 19\% & Szewczyk played 37 times for Poland, scoring & (Szewczyk, member of sports team, Poland, number of matches played, 37) \\
    & & 3 goals. & (Szewczyk, member of sports team, Poland, number of points, 3) \\
    \midrule
    Role & 12\% & John Sculley is a former Apple CEO. & (John Sculley, employer, Apple, position held, CEO) \\
    \midrule
    Part-Whole & 11\% & The Ohio Senate is the upper house of the Ohio & (Ohio, legislative body, Ohio General Assembly, has part, Ohio Senate) \\
    & & General Assembly, the Ohio state legislature. \\
    \midrule
    Location & 9\% & Donner was elected at the 1931 election as & (Donner, candidacy in election, 1931 election, electoral district, Islington West) \\
    & & Conservative MP for Islington West. \\
    \bottomrule
    \end{tabular}
    }
    \vspace{-1mm}
   \caption{General typology and distribution of frequent qualifier labels for the \dataname{} dataset, shown with example sentences and the corresponding hyper-relational facts.\vspace{-5mm}} 
    \label{tab:data_types}
\end{table*}

\subsection{Distantly Supervised Data Collection}
\label{sec:distant}
To collect a large and diverse dataset of sentences with hyper-relational facts, we employ distant supervision which falls under the weakly supervised setting.
Distant supervision automatically collects a dataset of relational facts by aligning a text corpus with facts from an existing knowledge graph.
Similar to \citet{elsahar-etal-2018-rex}, we first extract and link entities from the corpus to an existing knowledge graph, and resolve any coreference cases to the previously linked entities. 
To align hyper-relational facts from the knowledge graph to the text corpus, we detect if the entities that comprise each fact are also present in each sentence.
Each sentence with aligned facts is collected as part of the distantly supervised dataset.
To ensure that the large-scale text corpus can be well-aligned with the knowledge graph, we perform distant supervision between English Wikipedia and Wikidata \citep{DBLP:conf/semweb/ErxlebenGKMV14}, which is the central knowledge graph for Wikipedia. 
Following \citet{elsahar-etal-2018-rex}, we use the introduction sections of Wikipedia articles as the text corpus as they generally contain the most important information.

\paragraph{Entity Extraction and Linking}
The distant supervision process relies on matching entities in a sentence with facts from the knowledge graph.
To detect and identify the named entities in the articles, we use the DBpedia Spotlight \citep{10.1145/2063518.2063519} entity linker.
For the extraction of temporal and numerical entities, we use the spaCy \footnote{\href{https://spacy.io}{https://spacy.io}} tool.

\paragraph{Coreference Resolution}
As Wikipedia articles often use pronouns to refer to entities across sentences, it is necessary to resolve such references.
We employ the Stanford CoreNLP tool \citep{manning-etal-2014-stanford} for this task.

\paragraph{Hyper-Relational Alignment}
To extend the distant supervision paradigm to hyper-relational facts, we jointly match based on the entities that comprise each hyper-relational fact.
Formally, let $f = (e_{head}, r, e_{tail}, q, e_{value})$ be a possible hyper-relation fact consisting of the head entity, relation, tail entity, qualifier label and value entity, respectively.
Given a corpus of text articles, each article contains a set of sentences $\{s_i, ..., s_n\}$, where each sentence $s_i$ has $E_i$ entities that are linked to the knowledge graph.
For each hyper-relational fact $f$ in the knowledge graph, it is aligned to the sentence $s_i$ if the head entity $e_{head}$, tail entity $e_{tail}$ and value entity $e_{value}$ are all linked in the sentence.
Hence, we obtain a set of aligned facts for each sentence:  $\{(s_i, f) \mid e_{head} \in E_i, e_{tail} \in E_i, e_{value} \in E_i\}$.
Following \citet{10.1007/978-3-642-15939-8_10}, we remove any sentence that does not contain aligned facts.

\subsection{Human-Annotated Data Collection}
\label{sec:human}
Although distant supervision can align a large amount of hyper-relational facts, the process can introduce noise in the dataset due to possible spurious alignments and incompleteness of the knowledge graph \citep{7358050}.
However, it is not feasible to completely eliminate such noise from the dataset due to the annotation time and budget constraints.
Hence, we select a portion of the distantly supervised data to be manually labeled by human annotators.
To provide a solid evaluation setting for future research works, the human-annotated data will be used as the development and testing set.
We include the development set in the annotated portion as it is necessary for hyperparameter tuning and model selection.

The goal of the human annotation stage is to identify correct alignments and remove invalid alignments.
During the process, the annotators are tasked to review the correctness of each aligned fact, where an aligned fact consists of the sentence $s_i$ and hyper-relational fact $f$.
The alignment may be invalid if the relation triplet of the fact is not semantically expressed in the sentence, based on the Wikidata relation meaning.
For instance, given the sentence ``Prince Koreyasu was the son of Prince Munetaka who was the sixth shogun.'', the relation triplet (Prince Koreyasu, Occupation, shogun) is considered invalid as the sentence did not explicitly state if ``Prince Koreyasu'' became a shogun.
Similarly, the alignment may be invalid if the qualifier of the fact is not semantically expressed in the sentence, based on the Wikidata definition of the qualifier label.
For example, given the sentence ``Robin Johns left Northamptonshire at the end of the 1971 season.'', the hyper-relational fact (Robin Johns, member of sports team, Northamptonshire, Start Time, 1971) has an invalid qualifier as the label should be changed to ``End Time''.
Hence, the annotation is posed as a multi-class classification over each alignment with three classes: ``correct'', ``invalid triplet'' or ``invalid qualifier''.
Appendix \ref{sec:guide} has the annotation guide and data samples.

Each alignment sample is annotated by two professional annotators working independently.
There are 6780 sentences annotated in total and the inter-annotator agreement is measured using Cohen's kappa with a value of 0.56.
The kappa value is comparable with previous relation extraction datasets \citep{zhang-etal-2017-position}, demonstrating that the annotations are of reasonably high quality.
For each sample with disagreement, a third annotator is brought to judge the final result. 
We observe that 76\% of samples are annotated as ``correct'', which indicates a reasonable level of accuracy in the distantly supervised data.
To reduce the long-tailed class imbalance \citep{zhang-etal-2019-long}, we use a filter to ensure that all relation and qualifier labels have at least ten occurrences in the dataset.
\flag{
Although it can be more realistic to include challenging samples such as long-tailed class samples or negative samples in the dataset, we aim to address such challenges in a future dataset version release.
}


\begin{table}[!t]
    \centering
    \resizebox{1\columnwidth}{!}{
    \begin{tabular}{lrrrrcc}
    \toprule
    Dataset & \#Train & \#Dev & \#Test & \#Facts & $|R|$ & $|Q|$ \\
    \midrule
    TACRED  & 37,311 & 10,233 & 6,277 & 68,586 & 41 & 0  \\ 
    NYT24 & 56,196 & 5,000 & 5,000 & 17,624 & 24 & 0 \\
    NYT29 & 63,306 & 7,033 & 4,006 & 18,479 & 29 & 0 \\
    \dataname{} & 39,840 & 1,000 & 4,000 & 44,372 & 62 & 44 \\
    \bottomrule
    \end{tabular}
    }\vspace{-2mm}
   \caption{{Comparison of existing sentence-level datasets with \dataname{}.
   ``\#Fact'' denotes the unique facts, $|R|$ and $|Q|$ denote the unique relation labels and qualifier labels, respectively.
   To our knowledge, \dataname{} is the first RE dataset to include hyper-relational facts.
   }\vspace{-4mm}} 
    \label{tab:data_stats}
\end{table}

\subsection{Data Analysis}
\label{sec:data_analysis}

To provide a better understanding of the \dataname{} dataset, we analyze several aspects of the dataset.

\paragraph{Qualifier Typology}
The qualifiers of the hyper-relational facts can be grouped into several broad categories as shown in Table \ref{tab:data_types}.
Notably, the majority of the qualifiers fall under the ``Time'' category, as it can be considered a fundamental attribute of many facts.
The remaining qualifiers are distributed among the ``Quantity'', ``Role'', ``Part-Whole'' and ``Location'' categories.
Hence, the \dataname{} dataset is able to support a diverse typology of hyper-relational facts.

\paragraph{Size and Coverage}
The statistics of \dataname{} are shown in Table \ref{tab:data_stats}. 
We find that in terms of size and number of relation types, \dataname{} is comparable to existing sentence-level datasets, such as TACRED \citep{zhang-etal-2017-position}, NYT24 and NYT29 \citep{Nayak_Ng_2020}.
Table \ref{tab:data_types} also demonstrates that \dataname{} can serve as a general-purpose dataset, covering several domains such as business, sports and politics. Appendix \ref{sec:data_details} has more details.

\section{{\modelname{}: A Cube-Filling Approach}}

\subsection{Task Formulation}
\paragraph{Hyper-Relational Extraction}
Given an input sentence of $n$ words $s=\{x_1, x_2, ..., x_n\}$, an entity $e$ is a consecutive span of words where $e = \{x_i, x_{i+1}, ..., x_j\}, i,j \in \{1, ..., n\}$.
For each sentence $s$, the output of a hyper-relational extraction model is a set of facts where each fact consists of a relation triplet with an attributed qualifier.
A relation triplet consists of the relation $r \in R$ between head entity $e_{head}$ and tail entity $e_{tail}$ where $R$ is the predefined set of relation labels. 
The qualifier is an attribute of the relation triplet and is composed of the qualifier label $q \in Q$ and the value entity $e_{value}$, where $Q$ is the predefined set of qualifier labels.
Hence, a hyper-relational fact has five components: $(e_{head}, r, e_{tail}, q, e_{value})$.

\paragraph{Cube-Filling}
Inspired by table-filling approaches which can naturally perform binary relation extraction in an end-to-end fashion, we cast hyper-relational extraction as a cube-filling problem, as shown in Figure \ref{fig:cube}.
The cube contains multiple planes where the front-most plane is a two-dimensional table 
containing the entity and relation label information, while the following planes contain the corresponding qualifier information.
Each entry on the table diagonal represents a possible entity, while each entry outside the table diagonal represents a possible relation triplet.
For example, the entry ``Educated At'' represents a relation between the head entity ``Parker'' and the tail entity ``Harvard''.
Each table entry $y_{ij}^t$ can contain the null label $\bot$, an entity or relation label, i.e., $y_{ij}^t \in Y^t = \{\bot, \text{Entity}\} \cup{R}$.

The following planes in the cube represent the qualifier dimension, where each entry represents a possible qualifier label and value entity word for the corresponding relation triplet.
For instance, the entry ``Academic Degree'' in the qualifier plane for ``PhD'' corresponds to the relation triplet (Parker, Educated At, Harvard), hence forming the hyper-relational fact (Parker, Educated At, Harvard, Academic Degree, PhD).
Each qualifier entry $y_{ijk}^q$ can contain the null label $\bot$ or a qualifier label, i.e., $y_{ijk}^q \in Y^q = \{\bot\} \cup Q$.
Note that the cube-filling formulation also supports hyper-relational facts that share the same relation triplet, as the different qualifiers can occupy separate planes in the qualifier dimension and still correspond to the same relation triplet entry.


\subsection{Model Architecture}

Our model known as \modelname{} first encodes each input sentence using a language model encoder to obtain the contextualized sequence representation.
We then capture the interaction between each possible head and tail entity as a pair representation for predicting the entity-relation label scores.
To reduce the computational cost, each sentence is pruned to retain only words that have higher entity scores.
Finally, we capture the interaction between each possible relation triplet and qualifier 
to predict the qualifier label scores and decode the outputs.

\subsubsection{Sentence Encoding}

To encode a contextualized representation for each word in a sentence $s$, we use the pre-trained BERT \citep{devlin-etal-2019-bert} language model:
\begin{align}
    \{h_1, h_2, ..., h_n\} = \text{BERT}(\{x_1, x_2, ..., x_n\})
\end{align}
where $h_i$ denotes the contextualized representation of the i-th word in the sentence.

\begin{figure}[!t]
\centering
\includegraphics[width=1.0\columnwidth]{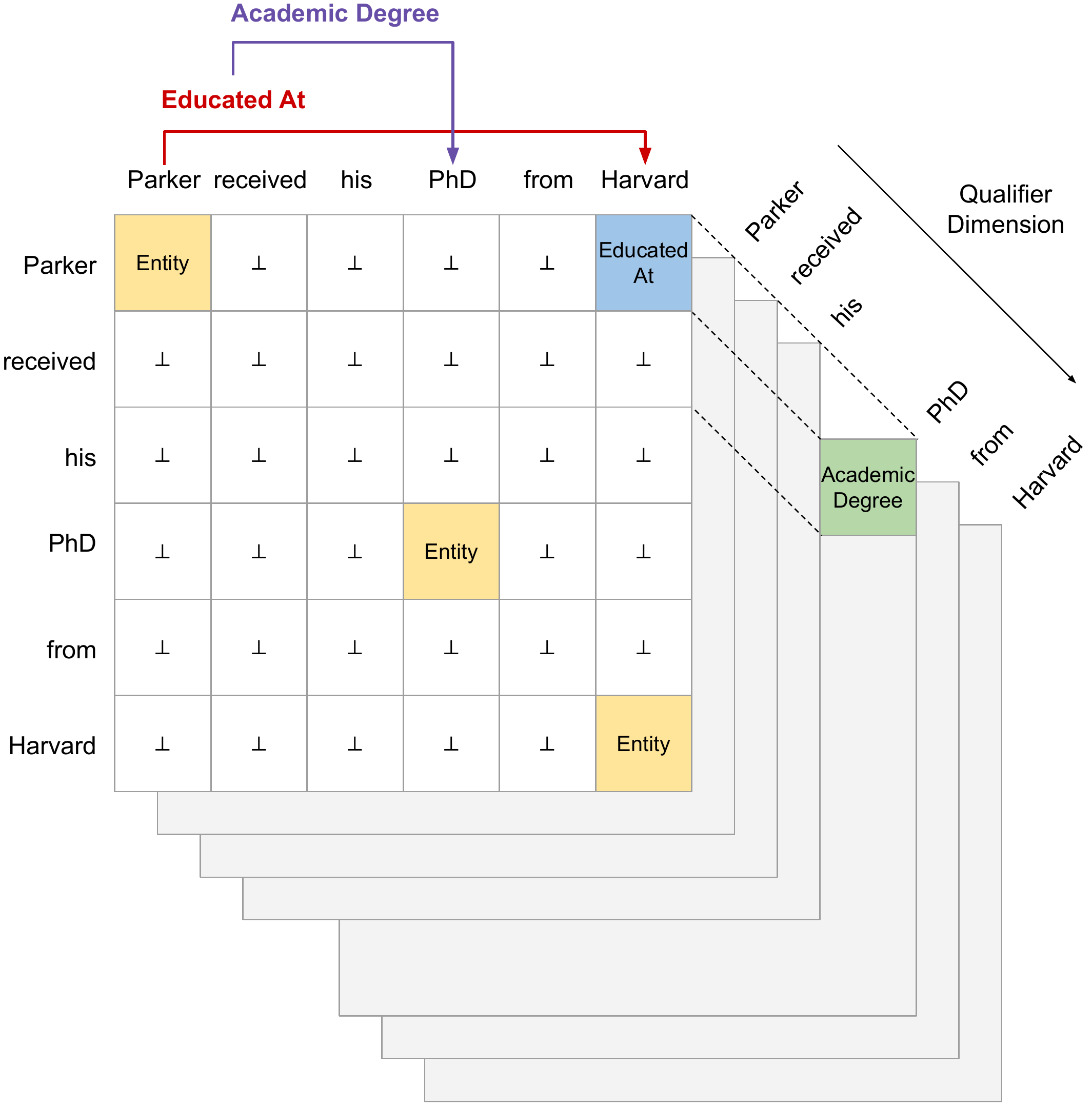}
\vspace{-3mm}
\caption{
An example of cube-filling for hyper-relational extraction.
The front-most plane is a two-dimensional table that contains entity and relation information.
It extends to the third dimension where each plane represents a possible qualifier label and value entity word that corresponds to the relation triplet entry.
\vspace{-5mm}}
\label{fig:cube}
\end{figure}

\subsubsection{Entity-Relation Representation}
To capture the interaction between head and tail entities, we concatenate each possible pair of word representations and project with a dimension-reducing feed-forward network (FFN): 
\begin{align} \label{eq:pair}
    g_{ij} = \text{FFN}_{pair}(h_i \oplus h_j)
\end{align}
Thus, we construct the table of categorical probabilities over entity and relation labels by applying an FFN and softmax over the pair representation:
\begin{align} \label{eq:r_score}
    P(\hat{y}_{ij}^t) = \text{Softmax}(\text{FFN}_{t}(g_{ij}))
\end{align}
where $\hat{y}_{ij}^t$ denotes the predicted table entry corresponding to the relation between the i-th possible head entity word and j-th possible tail entity word.
\flag{
Note that we use the concatenation operation in Equation \ref{eq:pair} instead of the averaging operation or other representation methods \citep{baldini-soares-etal-2019-matching} as the concatenation operation is simple and shown to be effective in recent RE works \citep{wang-etal-2021-unire, wang-lu-2020-two}.
}
\subsubsection{Cube-Pruning}
\label{sec:pruning}
To predict the qualifier of a hyper-relational fact, the model needs to consider the interaction between each possible relation triplet and value entity, where the relation triplet contains a head entity and a tail entity.
For a sentence with $n$ words, there are $n^3$ interactions that do not scale well for longer input sequences. 
Hence, we propose a cube-pruning method to consider only interactions between the top $m$ words in terms of entity score.
Consequently, the model will only consider the interaction between the top-$m$ most probable words of the potential head entities, tail entities, and value entities respectively.
This reduces the number of interactions to $m^3$ where $m$ is a fixed hyperparameter.
The cube-pruning method also has the benefit of alleviating the negative class imbalance by reducing the proportion of entries with the null label, and we analyze this effect in Section \ref{sec:pruning_effect}.
To detect the most probable entity words, we obtain the respective entity scores from the diagonal of the table $\hat{y}^t$ containing the entity and relation scores (i.e., the front-most plane in Figure \ref{fig:cube}):
\begin{align}
    \Phi_i^{entity} = P(\hat{y}_{ii}^{t}), i \in \{1, ..., n\}
\end{align}
The entity scores are then ranked to obtain the pruned indices $\{1, ..., m\}$ which will be applied to each dimension of the cube representation. 

To capture the hyper-relational structure between relation triplets and qualifier attributes, we use a bilinear interaction layer between each possible pair representation and word representation.
The categorical probability distribution over qualifier labels for each possible relation triplet and value entity is then computed as:
\begin{align} \label{eq:q_score}
    P(\hat{y}^q_{i'j'k'}) = \text{Softmax}(g_{i'j'}^\intercal \; U \; h_{k'})
\end{align}
where $i',j',k' \in \{1,...,m\}$ are the pruned indices and $U$ is a trainable bilinear weight matrix.

\subsubsection{Training Objective}
The training objective for the entity-relation table is computed using the negative log-likelihood as:
\begin{align}
    \mathcal{L}_{t} = - \frac{1}{n^2} \sum_{i=1}^{n} \sum_{j=1}^{n} \text{log} P(
    \hat{y}_{ij}^t)
\end{align}
The training objective for the qualifier dimension is computed using the negative log-likelihood as:
\begin{align}
    \mathcal{L}_{q} = - \frac{1}{m^3} \sum_{i'=1}^{m} \sum_{j'=1}^{m} \sum_{k'=1}^{m} \text{log} P(
    \hat{y}_{i'j'k'}^q)
\end{align}
To enable end-to-end training, the overall cube-filling objective is aggregated as the sum of losses:
\begin{align}
    \mathcal{L} = \mathcal{L}_{t} + \mathcal{L}_{q}
\end{align}

\subsubsection{Decoding}
To decode the hyper-relational facts from the predicted scores, we implement a simple and efficient method and provide the pseudocode in Appendix \ref{sec:decode_algo}.
As it is intractable to consider all possible solutions, a slight drop in decoding accuracy is acceptable.
A key intuition is that if a valid qualifier exists, this indicates that a corresponding relation triplet also exists.
Hence, we first decode the qualifier scores 
(Equation \ref{eq:q_score})
to determine the span positions of the head entity, tail entity and value entity in each hyper-relational fact.
Consequently, we can determine the relation and qualifier label from the corresponding entries in the relation scores (Equation \ref{eq:r_score}) and qualifier scores respectively.

To handle entities that may contain multiple words, we consider adjacent non-null qualifier entries to correspond to the same head entity, tail entity, and value entity, hence belonging to the same hyper-relational fact. 
This assumption holds true for 97.14\% of facts in the dataset.
To find and merge the adjacent non-null entries, we use the nonzero operation which is more computationally efficient compared to nested for-loops.
For each group of adjacent entries that correspond to the same hyper-relational fact, we determine the relation label by averaging the corresponding relation scores.
Similarly, we determine the qualifier label by averaging the corresponding qualifier scores.
When using cube-pruning, we map the pruned indices back to the original indices before decoding.
Appendix \ref{sec:efficiency} has the model speed comparison.


\section{Experiments}

\subsection{Experimental Settings}

\paragraph{Evaluation} 
Similar to other information extraction tasks, we use the Micro $F_1$ metric for evaluation on the development and test set.
For a predicted hyper-relational fact to be considered correct, the whole fact $f = (e_{head}, r, e_{tail}, q, e_{value})$ must match the ground-truth fact in terms of relation label, qualifier label and entity bounds.

\paragraph{Hyperparameters} 
For the encoding module, we use the BERT language model, specifically the uncased base and large versions. 
We train for 30 epochs with a linear warmup for 20\% of training steps and a maximum learning rate of 5e-5. 
We employ AdamW as the optimizer and use a batch size of 32. 
For model selection and hyperparameter selection, we evaluate based on the $F_1$ on the development set.
We use $m=20$ for cube-pruning and Appendix \ref{sec:hparams} has more experimental details.

\begin{table*}[!t]
    \centering
    \resizebox{1\textwidth}{!}{
    \begin{tabular}{lccccccc}
    \toprule
    \multirow{2}{*}{{\textbf{Model}}} & \multirow{2}{*}{{\textbf{Parameters}}} 
    & & \textbf{Dev} & & & \textbf{Test} & \\
    \cmidrule(lr){3-5}
    \cmidrule(lr){6-8}
    & & Precision & Recall & $F_1$ & Precision & Recall & $F_1$ \\
    \midrule
    Generative Baseline (Base) & 140M 
    & 63.79 $\pm$ 0.27 & 59.94 $\pm$ 0.68 & 61.80 $\pm$ 0.37
    & 64.60 $\pm$ 0.47 & 59.67 $\pm$ 0.35 & 62.03 $\pm$ 0.21 \\
    Pipeline Baseline (Base) & 132M 
    & \textbf{69.23} $\pm$ 0.30 & 58.21 $\pm$ 0.57 & 63.24 $\pm$ 0.44
    & \textbf{69.00} $\pm$ 0.48 & 57.55 $\pm$ 0.19 & 62.75 $\pm$ 0.29 \\
    \modelname{} (Base) & 115M 
    & 66.14 $\pm$ 0.88 & \textbf{64.39} $\pm$ 1.23 & \textbf{65.24} $\pm$ 0.82 
    & 65.82 $\pm$ 0.84 & \textbf{64.28} $\pm$ 0.25 & \textbf{65.04} $\pm$ 0.29 \\
    \midrule
    {Generative Baseline (Large)} & 400M
    & 67.08 $\pm$ 0.49 & 65.73 $\pm$ 0.78 & 66.40 $\pm$ 0.47
    & \textbf{67.17} $\pm$ 0.40 & 64.56 $\pm$ 0.58 & 65.84 $\pm$ 0.25 \\
    {\modelname{} (Large)} & 343M
    & \textbf{68.75} $\pm$ 0.82 & \textbf{68.88} $\pm$ 1.03 & \textbf{68.81} $\pm$ 0.46
    & 66.39	$\pm$ 0.96 & \textbf{67.12} $\pm$ 0.69 & \textbf{66.75} $\pm$ 0.65 \\
    \bottomrule
    \end{tabular}
    }
    \vspace{-2mm}
    \caption{Evaluation results for hyper-relational extraction on the \dataname{} dataset.\vspace{-4mm}}
    \label{tab:results}
\end{table*}


\subsection{Baseline Methods}
\label{sec:baseline}
As there are no existing models for hyper-relational extraction, we introduce two strong baselines that leverage pretrained language models. 
The pipeline baseline is based on a competitive table-filling model for joint entity and relation extraction, while the generative baseline is extended from a state-of-the-art approach for end-to-end relation extraction.

\paragraph{Pipeline Baseline}
As pipeline methods can serve as strong baselines for information extraction tasks \citep{zhong-chen-2021-frustratingly}, we implement a pipeline method for hyper-relational extraction.
Concretely, we first train a competitive relation extraction model architecture UniRE \citep{wang-etal-2021-unire} to extract relation triplets from each input sentence.
Separately, we train a span extraction model based on BERT-Tagger \citep{devlin-etal-2019-bert} that is conditioned on the input sentence and a relation triplet to extract the value entities and corresponding qualifier label.
However, as both stages fine-tune a pretrained language model, the pipeline method doubles the number of trainable parameters compared to an end-to-end method which only fine-tunes one pretrained language model.
To avoid an unfair comparison as larger models are more sample-efficient \citep{DBLP:journals/corr/abs-2001-08361}, we use DistilBERT \citep{DBLP:journals/corr/abs-1910-01108} in both stages of the pipeline.

\paragraph{Generative Baseline}
Inspired by the flexibility of language models for complex tasks such as information extraction and controllable structure generation \citep{shen-etal-2022-mred}, we propose a generative method for hyper-relational extraction. 
Compared to a pipeline method, a generative method can perform hyper-relational extraction in an end-to-end fashion without task-specific modules \citep{paolini2021structured}.
Similar to existing generative methods for relation extraction \citep{huguet-cabot-navigli-2021-rebel-relation, chia-etal-2022-relationprompt}, we use BART \citep{lewis-etal-2020-bart} which takes the sentence as input and outputs a structured text sequence that is then decoded to form the extracted facts.
For instance, given the sentence ``Parker received his PhD from Harvard.'', the sequence-to-sequence model is trained to generate ``Head Entity: Parker, Relation: educated at, Tail Entity: Harvard, Qualifier: academic degree, Value: PhD.''
The generated text is then decoded through simple text processing to form the hyper-relational fact (Parker, Educated At, Harvard, Academic Degree, PhD). 

\subsection{Main Results}
\label{sec:main_results}
{
We compare CubeRE with the baseline models and report the precision, recall, and $F_1$ scores with standard deviation in Table \ref{tab:results}.
The results demonstrate the general effectiveness of our model as \modelname{} has consistently higher $F_1$ scores on both the base and large model settings.
While the pipeline baseline relies on a two-stage approach that is prone to error propagation, \modelname{} can perform hyper-relational extraction in an end-to-end fashion. 
Hence, \modelname{} is able to detect more valid hyper-relational facts, which is demonstrated by the higher recall and $F_1$ scores.
}
Compared to the generative baseline, our cube-filling approach is able to explicitly consider the interaction between relation triplets and qualifiers to better extract hyper-relational facts.
Furthermore, we argue that \modelname{} is more interpretable than the generative baseline as it can compute the score for each possible relation triplet and qualifier.
Hence, \modelname{} can also be more controllable as 
it is possible to control the number of predicted facts by applying a threshold to the triplet and qualifier scores.

\begin{table}[!t]
    \centering
    \resizebox{1\columnwidth}{!}{
    \begin{tabular}{lccc}
    \toprule
    Model & Precision & Recall & $F_1$ \\
    \midrule
    Generative Baseline & 69.96	$\pm$ 0.31 & 64.56	$\pm$ 0.21 & 67.15 $\pm$ 0.09 \\
    Pipeline Baseline & 75.94 $\pm$ 0.66 & 66.41 $\pm$ 0.72 & 70.85 $\pm$ 0.13 \\
    \modelname{} & 72.45 $\pm$ 0.66 & 69.64 $\pm$ 0.53 & 71.01 $\pm$ 0.16 \\
    \bottomrule
    \end{tabular}
    }
   \caption{Evaluation results on \dataname{} considering only the triplet component of hyper-relational facts.} 
    \label{tab:triplet_results}
\end{table}

\subsection{Triplet-Based Evaluation}
{
To further investigate the differences in model performance, we also report the results when considering only the triplet component of hyper-relational facts in Table \ref{tab:triplet_results}.
The results show that \modelname{} has comparable performance to the pipeline baseline when considering only relation triplets.
Hence, this suggests that the performance improvement in hyper-relational extraction is most likely due to more accurate qualifier extraction. 
Compared to the pipeline baseline which has two separate encoders for triplet extraction and conditional qualifier extraction, \modelname{} learns a shared representation of the input sentence that is guided by both the triplet and qualifier losses facilitating the interaction between relation triplets and qualifiers.
The triplet-qualifier interaction is important as most qualifier labels are relatively relation-specific\footnote{Please refer to Appendix \ref{sec:data_details} for the qualifier analysis.}.
This allows \modelname{} to extract the qualifiers more accurately, resulting in better overall performance.
}

\section{Analysis}
\label{sec:analysis}
In this section, we study the effect of cube-pruning and identify directions for future research. Further analysis is shown in Appendix \ref{sec:further}.

\begin{figure}
\centering
\resizebox{0.75\linewidth}{!}{
\begin{tikzpicture}
\pgfplotsset{width = 6cm, height = 4cm}
    \begin{axis}[
        ymax=67,
        ymin=61,
        ylabel={$F_1$ (\%)},
        label style={font=\fontsize{7}{1}\selectfont},
        xtick = {1,2,3,4,5},
        xticklabels = {10, 20, 30, 40, $\infty$},
        xticklabel style = {font=\fontsize{7}{1}\selectfont},
        yticklabel style = {font=\fontsize{7}{1}\selectfont},
        xtick pos = left,
        ytick pos = left,
        ymajorgrids = true,
        grid style=dashed,
    ]
    \addplot [mark=square, mark size=1.2pt, color=orange] plot coordinates {
    (1, 64.24) (2, 65.77) (3, 64.33) (4, 63.60) (5, 62.35)};
    \end{axis}
\end{tikzpicture}
}
\vspace{-3mm}
\caption{The effect of pruning threshold $m$ on Dev $F_1$. The model without pruning is indicated as $m=\infty$.\vspace{-5mm}}
\label{fig:prune}
\end{figure}
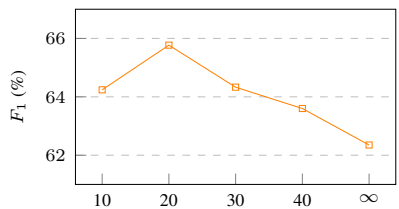

\subsection{Effect of Pruning}
\label{sec:pruning_effect}

In addition to improving the computational efficiency of \modelname{} as discussed in Section \ref{sec:pruning}, our cube pruning method may also improve the extraction performance of the model. 
During training, the cube-filling approach faces the issue of having mostly null entries, thus biasing the learning process with negative class imbalance \cite{li-etal-2020-dice}.
By pruning the cube to consider only the entries associated with higher entity scores, the proportion of null entries is reduced, hence alleviating the class imbalance issue. 
This is supported by the trend in Figure \ref{fig:prune}, as relaxing the pruning threshold $m$ leads to reduced $F_1$ scores.
On the other hand, overly strict pruning will reduce the recall, negatively affecting the overall performance.

\subsection{Model Performance Breakdown}
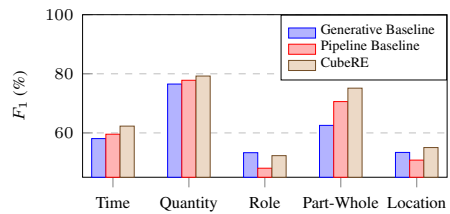
\begin{figure}[t!]
\centering
\resizebox{0.85\linewidth}{!}{
\begin{tikzpicture}
\pgfplotsset{width = 7cm, height = 4cm}
    \begin{axis}[
        ybar=0cm,
        bar width=6pt,
        ymax=100,
        ymin=45,
        ylabel={$F_1$ (\%)},
        label style={font=\fontsize{7}{1}\selectfont},
        xtick = {1,2,3,4,5},
        xticklabels = {Time, Quantity, Role, Part-Whole, Location},
        xticklabel style = {font=\fontsize{7}{1}\selectfont},
        yticklabel style = {font=\fontsize{7}{1}\selectfont},
        xtick pos = left,
        ytick pos = left,
        ymajorgrids = true,
        grid style=dashed, 
        legend style={
            font=\fontsize{6}{1}\selectfont, 
            legend style={row sep=-0.1cm},
            at={(1,1)},
        },
        legend image code/.code={
          \draw[#1] (0cm,-0.1cm) rectangle (0.4cm,0.05cm);
        }, 
        legend cell align={left},
    ]
    \addplot coordinates {
    (1, 58.08) (2, 76.53) (3, 53.31) (4, 62.55) (5, 53.40)};
    \addlegendentry{Generative Baseline};
    \addplot coordinates {
    (1, 59.56) (2, 77.81) (3, 48.06) (4, 70.60) (5, 50.80)};
    \addlegendentry{Pipeline Baseline};
    \addplot coordinates {
    (1, 62.30) (2, 79.24) (3, 52.31) (4, 75.14) (5, 55.05)};
    \addlegendentry{\modelname{}};
    \end{axis}
\end{tikzpicture}
\vspace{-2mm}
}
\caption{{
Model performance breakdown based on the general categories of qualifiers as shown in Table \ref{tab:data_types}.
\vspace{-3mm}}
}
\label{fig:breakdown}
\end{figure}

To identify directions for future research in hyper-relational extraction, we analyze the model performance separately for each general qualifier category.
As shown in Table \ref{fig:breakdown}, there is a variance in model performance across qualifier categories that cannot be fully explained by their proportion in the dataset.
For instance, although the ``Time'' category comprises a majority of the qualifiers, it does not have the highest performance.
This suggests that future research may focus on areas such as temporal reasoning, which is an open challenge for language models \citep{vashishtha-etal-2020-temporal, dhingra-etal-2022-time}.
In addition, \modelname{} demonstrates strong performance across all categories which suggests that it can serve as a general extraction model for different qualifiers.

\section{Related Work}

\paragraph{Knowledge Graph Construction}

In addition to extraction from natural language text, the underlying facts for knowledge graphs can also be extracted from semi-structured websites \citep{10.14778/3231751.3231758}, tables \citep{dong-etal-2020-multi-modal} or link prediction \citep{8047276}.
However, textual extraction may be a more pressing challenge due to the vast amount of unstructured textual data on the web \citep{10.1145/3336191.3371878}. 
Hence, this work focuses on extracting facts from unstructured text.

\paragraph{Relation Extraction}

Although relation extraction is a well-established task, most methods only consider the relation between two entities.
There have been several directions to extract more complex facts, such as n-ary relation extraction or document-level relation extraction \citep{yao-etal-2019-docred}.
However, n-ary relation extraction \citep{jia-etal-2019-document, akimoto-etal-2019-cross} has a limited scope as the available datasets address the biomedical domain.
On the other hand, document-level \citep{tan-etal-2022-document} and cross-document relation extraction \citep{yao-etal-2021-codred} are fundamentally limited by the binary relation structure which does not consider hyper-relational information.
\flag{
Although dialogue-level relation extraction \citep{DBLP:journals/corr/abs-2009-05092} may have a more complex structure consisting of utterances and speaker information, current datasets \citep{welleck-etal-2019-dialogue} focus on the binary relation format.
}
Hence, we propose to fill the gap by contributing \dataname{}, a general-purpose and large-scale dataset for hyper-relational extraction that is not limited to any specific domain. 

\paragraph{Information Extraction}
\flag{
In this work, we focus on relation extraction which falls under the broad scope of information extraction \citep{bing-etal-2015-improving}.
Hence, a possible future direction is to adapt \modelname{} for extracting other types of information such as attributes \citep{10.1145/2433396.2433468}, events \citep{wang-etal-2021-cleve}, arguments \citep{cheng-etal-2020-ape, cheng-etal-2022-iam}, aspect-based sentiment \citep{xu-etal-2021-learning, 10.1145/3459637.3482058}, commonsense knowledge \citep{ghosal-etal-2021-cider}, or visual scene relations \citep{DBLP:journals/corr/abs-1909-06273}.
Additionally, as \dataname{} relies on distant supervision for dataset construction, it is necessary to further explore how to mitigate the noise in distantly supervised datasets for information extraction tasks \citep{nayak-etal-2021-improving}.
}

\paragraph{Table-Filling}
Table-Filling is a popular approach for entity and relation extraction tasks \citep{miwa-sasaki-2014-modeling, gupta-etal-2016-table, zhang-etal-2017-position}.
It has several advantages including interpretability and an end-to-end formulation.
Hence, table-filling approaches are able to avoid the cascading error propagation faced by pipeline models, despite a compact parameter set.
Inspired by the benefits of table-filling, we extend the approach to cube-filling to extract hyper-relational facts by considering qualifiers for each relation triplet.
To our knowledge, our proposed model is the first cube-filling approach for information extraction tasks.

\section{Conclusions}

In this work, we propose the hyper-relational extraction task for extracting richer and more complete facts from natural text. 
To support the task, we introduce \dataname{}, a large-scale and general-purpose dataset that is not restricted to any specific domain.
As there is no available model for hyper-relational extraction, we propose an end-to-end cube-filling approach
inspired by table-filling methods for relation extraction.
We further propose a cube-pruning method to reduce computational cost and alleviate negative class imbalance during training.
Experiments on \dataname{} demonstrate the effectiveness of \modelname{} compared to strong baselines, setting the benchmark for future work.

\section*{Limitations}

\paragraph{Model Limitations}
Regarding the \modelname{} model, we propose a cube-pruning method to improve the computational efficiency and reduce the negative class imbalance. 
The cube-pruning threshold is fixed, although the input can have different sentence lengths.
Hence, it may result in overly strict pruning if the sentence is extremely long. 
However, the pruning threshold is similar to the maximum sequence length in most transformer-based models and may need to be tuned according to the specific dataset or application scenario.
The optimal cube-pruning threshold is selected based on the analysis in Section \ref{sec:pruning_effect}.
\modelname{} may not work well for overlapping or nested entity spans, which affects 2.11\% of the sentences.
This can be considered a general limitation of table-filling methods for relation extraction, and future work may need to consider a span-based approach \cite{xu-etal-2021-learning} to address this issue.
\paragraph{Data Limitations} 
Regarding the \dataname{} dataset, the distant supervision method of data collection may not align all valid facts present in the text articles.
This is due to the possible incompleteness of the knowledge graph which is an open research challenge \citep{7358050}.
On the other hand, it is not feasible to manually annotate all possible facts due to constraints in annotation time and cost. 
Furthermore, there are a large number of relation and qualifier labels to consider, resulting in a challenging task for human annotators.
A promising and practical method to address the challenges in distant supervision is to adopt a human-in-the-loop annotation scheme for RE \citep{DBLP:journals/corr/abs-2205-12696}.
The annotation scheme can increase the number of facts in a dataset by training a RE model to predict more candidate facts for each text article, which are then reviewed and filtered by humans.
However, this model-assisted annotation approach is not applicable to the construction of \dataname{} as it relies on existing strong RE models, whereas there are no suitable models for hyper-relational extraction existing prior to this work.


\section*{Ethics Statement}
\paragraph{Model Ethics}
Regarding the model generalization, we expect that the models introduced should perform similarly for factual text articles such as news articles from various domains, similar to the proposed dataset.
However, it may not perform well for more casual text formats such as chat discussions or opinion pieces.
On the other hand, we note that the models extract hyper-relational facts from the input sentences and do not guarantee the factual correctness of the extracted facts.
This is an ethical consideration of RE models in general and further fact verification \citep{Nie_Chen_Bansal_2019} modules are necessary before the facts can be integrated into knowledge graphs or downstream applications.
\paragraph{Data Ethics} 
For the dataset construction, we collect texts and facts from Wikipedia and Wikidata respectively, which is a common practice for distantly supervised datasets.
Wikidata facts are under the public domain\footnote{\href{https://www.wikidata.org/wiki/Wikidata:Licensing}{https://www.wikidata.org/wiki/Wikidata:Licensing}} while Wikipedia texts are licensed under the Creative Commons Attribution-ShareAlike 3.0 Unported License\footnote{\href{https://en.wikipedia.org/wiki/Wikipedia:Copyrights}{https://en.wikipedia.org/wiki/Wikipedia:Copyrights}}. 
Hence, we are free to adapt the texts to construct our dataset, which will also be released under the same license.
For the human data annotation stage, we employ two professional data annotators, and they have been fairly compensated.
The compensation is negotiated based on the task complexity and assessment of the reasonable annotation speed.
Based on the agreed annotation scheme, each annotation batch is required to undergo quality checking where a portion of samples are manually checked.
If any batch does not meet the acceptance criteria of 95\% accuracy, the annotators are required to fix the errors before the batch can be accepted.
The overall quality of the dataset is evaluated in Section \ref{sec:distant} and Section \ref{sec:human}, and we analyze the dataset characteristics in Section \ref{sec:data_analysis}, with further analysis in Section \ref{sec:size}.



\clearpage
\newpage

\bibliography{anthology,custom}
\bibliographystyle{acl_natbib}

\begin{table*}[t]
    \centering
    \resizebox{1\textwidth}{!}{
    \begin{tabular}{llrrrrr}
    \toprule
    Data Setting & Annotation Type & Sentences & Facts & Entities & Average Sentence Length & Average Entity Length \\
    \midrule
    Train & Distant-Supervised & 39,840 & 39,978 & 32,539 & 31.91 words & 1.67 words \\
    Dev & Human Annotated & 1,000 & 1,220 & 1,912 & 30.30 words & 1.71 words \\
    Test & Human Annotated & 4,000 & 4,796 & 5,842 & 30.06 words & 1.69 words \\
    \bottomrule
    \end{tabular}
    }
  \caption{Detailed statistics for the \dataname{} dataset.} 
    \label{tab:data_details}
\end{table*}

\appendix


\section{Annotation Guide}
\label{sec:guide}
This section explains the guideline for human annotators.
The task is a classification of whether each hyper-relational fact can be reasonably extracted from a piece of text.
Each annotation sample contains one sentence and one corresponding fact for judgment.
The annotator should classify each sample as “Correct” or “Invalid Triplet” or “Invalid Qualifier”.
Each hyper-relational fact has five components with the format (head entity, relation label, tail entity, qualifier label, value entity).
The head entity is the main subject entity of the relationship.
The relation label is the  category of relationship that is expressed between the head and tail entity.
The tail entity is the object entity of the relationship that is paired with the head entity.
The qualifier label is the category of the qualifier information.
The value entity is the corresponding value of the qualifier that is applied to the relation triplet (head, relation, tail).

The value entity can contain a date, quantity, or short piece of text which is the mentioned name of the entity.
For the annotation objective, we want to know whether this piece of information is clearly expressed by the given text.
All the entities, relations, and qualifiers exist in the Wikidata database, so annotators can refer to the relation or qualifier definition at https://www.wikidata.org for clarification.
The annotation steps are as follows:
\begin{enumerate}
    \item Read and understand the text sample which is a continuous sequence of words. 
    Then, consider the corresponding hyper-relational fact. 
    \item First check the triplet (head, relation, tail) of the fact.
    If the head and tail entity mentioned in the text do not clearly express the relation's meaning, then the whole fact should be marked as “Invalid Triplet”.
    \item Check the (qualifier, value) components. If the value mentioned in the text does not clearly express the qualifier meaning or is not directly related to the triplet, then the fact should be marked as “Invalid Qualifier”.
    \item If there is no error in the fact, then it can be marked as “Correct”.
\end{enumerate}

For example, given the sentence ``The film's story earned Leonard Spigelgass a nomination as Best Story for the 23rd Academy Awards.'', the fact (Leonard Spigelgass, nominated for, Best Story, statement is subject of, 23rd Academy Awards) is correct as Leonard was nominated and the main topic is the Academy Awards.
However, given the sentence ``Prince Koreyasu was the son of Prince Munetaka who was the sixth shogun.'', the fact (Prince Koreyasu, occupation, shogun, replaces, Prince Munetaka) has an invalid triplet as we don’t know if Koreyasu became a shogun.
On the other hand, given the sentence “Robin Johns left Northamptonshire at the end of 
the 1971 season.”, the fact (Robin 
Johns, member of sports team, Northamptonshire, 
Start Time, 1971) has an invalid qualifier as the 
qualifier label should be “End Time” instead of ``Start Time''.

\begin{table*}[!t]
    \centering
    \resizebox{1\textwidth}{!}{
    \begin{tabular}{lccccccc}
    \toprule
    \multirow{2}{*}{{\textbf{Model}}} & \multirow{2}{*}{{\textbf{Parameters}}} 
    & & \textbf{Dev} & & & \textbf{Test} & \\
    \cmidrule(lr){3-5}
    \cmidrule(lr){6-8}
    & & Precision & Recall & $F_1$ & Precision & Recall & $F_1$ \\
    \midrule
    Generative Baseline & 140M 
    & 63.79 $\pm$ 0.27 & 59.94 $\pm$ 0.68 & 61.80 $\pm$ 0.37
    & 64.60 $\pm$ 0.47 & 59.67 $\pm$ 0.35 & 62.03 $\pm$ 0.21 \\
    Pipeline Baseline & 132M 
    & \textbf{69.23} $\pm$ 0.30 & 58.21 $\pm$ 0.57 & 63.24 $\pm$ 0.44
    & \textbf{69.00} $\pm$ 0.48 & 57.55 $\pm$ 0.19 & 62.75 $\pm$ 0.29 \\
    \modelname{} & 115M 
    & 66.14 $\pm$ 0.88 & \textbf{64.39} $\pm$ 1.23 & \textbf{65.24} $\pm$ 0.82 
    & 65.82 $\pm$ 0.84 & \textbf{64.28} $\pm$ 0.25 & \textbf{65.04} $\pm$ 0.29 \\
    \midrule
    {Pipeline Baseline (Medium)} & 221M
    & 69.70 $\pm$ 1.08 & 62.33 $\pm$ 0.50 & 65.80 $\pm$ 0.54
    & 69.38	$\pm$ 0.39 & 61.96 $\pm$ 0.54 & 65.46 $\pm$ 0.32 \\
    \midrule
    {Generative Baseline (Large)} & 400M
    & 67.08 $\pm$ 0.49 & 65.73 $\pm$ 0.78 & 66.40 $\pm$ 0.47
    & \textbf{67.17} $\pm$ 0.40 & 64.56 $\pm$ 0.58 & 65.84 $\pm$ 0.25 \\
    {\modelname{} (Large)} & 343M
    & \textbf{68.75} $\pm$ 0.82 & \textbf{68.88} $\pm$ 1.03 & \textbf{68.81} $\pm$ 0.46
    & 66.39	$\pm$ 0.96 & \textbf{67.12} $\pm$ 0.69 & \textbf{66.75} $\pm$ 0.65 \\
    \midrule
    {\flag{Pipeline Baseline (Large)}} & 680M
    & 70.58	$\pm$ 0.78 & 66.58 $\pm$ 0.66 & 68.52 $\pm$ 0.32
    & 69.21 $\pm$ 0.55 & 64.27 $\pm$ 0.24 & 66.65 $\pm$ 0.28 \\
    \bottomrule
    \end{tabular}
    }
    \vspace{-2mm}
    \caption{Evaluation results for hyper-relational extraction on the HyperRED dataset.\vspace{-4mm}}
    \label{tab:results2}
\end{table*}

\begin{table}[!t]
    \centering
    \resizebox{1\columnwidth}{!}{
    \begin{tabular}{lrrr}
    \toprule
    Model & Training Time & Inference Speed & Memory Usage \\
    \midrule
    Generative & 1.93 hrs & 37 samples/s & 3.9 GB \\
    Pipeline & 2.41 hrs & 181 samples/s & 5.5 GB \\
    \modelname{} & 3.08 hrs & 160 samples/s & 6.6 GB \\
    \bottomrule
    \end{tabular}
    }
   \caption{Comparison of the computational cost for the Generative, Pipeline and \modelname{} models.} 
    \label{tab:costs}
\end{table}

\begin{table}[!t]
    \centering
    \resizebox{1\columnwidth}{!}{
    \begin{tabular}{lr}
    \toprule
    & Experimental Detail \\
    \midrule
    GPU Model & Nvidia V100 \\
    CUDA Version & 11.3 \\
    Python Version & 3.7.12 \\
    PyTorch Version & 1.11.0 \\
    Wikidata Version & 20170503 \\
    Long-Tailed Threshold & 10 \\
    Pruning Threshold & 20 \\
    Maximum Sequence Length (words) & 80 \\
    FFN Hidden Size & 150 \\
    Learning Rate Decay & 0.9 \\
    Adam Epsilon & 1e-12 \\
    Adam Weight Decay Rate & 1e-5 \\
    \bottomrule
    \end{tabular}
    }
   \caption{List of experimental details.} 
    \label{tab:hparams}
\end{table}

\section{Experiment Details}
\label{sec:hparams}
\paragraph{Hyperparameters}
Table \ref{tab:hparams} shows the details of our experimental setup and model hyperparameters.
For the analysis experiments in Section \ref{sec:analysis}, we use the BERT-Base version of \modelname{} and report the $F_1$ metric score on the development set of \dataname{} unless otherwise stated in the specific subsection.

\paragraph{Pipeline Baseline Details}
For the pipeline baseline, we use DistilBERT as the language model encoder for both the triplet extraction and conditional qualifier extraction stages.
Both stages of the pipeline are fine-tuned separately on the gold labels.
At inference time, the triplet extraction stage takes the sentence as input and outputs the predicted relation triplets.
For each predicted relation triplet, the conditional qualifier extractor takes the sentence and the relation triplet as input to predict the possible qualifiers where each qualifier consists of the qualifier label and value entity.
The input of the qualifier extraction model is the concatenated sentence and relation triplet.
For example, the sentence ``Leonard Parker received his PhD from Harvard University in 1967.'' and relation triplet (Leonard Parker, Educated At, Harvard University), will be concatenated to become `Leonard Parker received his PhD from Harvard University in 1967. Leonard Parker | Educated At | Harvard University''.
The outputs of both stages are then merged to form the predicted hyper-relational facts.
Following the BERT-Tagger, the conditional qualifier extraction model is trained using the crossentropy loss for sequence labeling.
To encode the qualifier information as sequence labels, we use the BIO tagging scheme where the sequence label corresponds to the possible qualifier label for each entity word.
For both stages which are trained separately, we use the same epochs, learning rate and batch size as the \modelname{} model for fairness.

\paragraph{Generative Baseline Details}
The generative baseline model can predict hyper-relational facts by learning to generate a text sequence with a special structured format as demonstrated in Section \ref{sec:baseline}.
Note that if the sentence contains multiple hyper-relational facts, the desired output sequence is simply the concatenated text sequence of the structured text for each fact.
The multiple facts can be easily decoded from the structured text format with simple text processing such as regex.
As the input and output of the model are text sequences which do not violate the model vocabulary, the generative baseline can be trained using a standard sequence-to-sequence modeling objective.
For training, we use the same epochs, learning rate and batch size as the \modelname{} model for fairness.

\begin{figure}[t!]
\centering
\resizebox{1.0\linewidth}{!}{
\begin{tikzpicture}
\pgfplotsset{width = 7cm, height = 4cm}
    \begin{axis}[
        ybar=0cm,
        bar width=10pt,
        ymax=35,
        ymin=0,
        label style={font=\fontsize{7}{1}\selectfont},
        xtick = {1,2,3,4},
        xticklabels = {1-5, 6-10, 11-15, 16+},
        xticklabel style = {font=\fontsize{7}{1}\selectfont},
        yticklabel style = {font=\fontsize{7}{1}\selectfont},
        xtick pos = left,
        ytick pos = left,
    ]
    \addplot coordinates {
    (1, 32) (2, 8) (3, 1) (4, 3)};
    \end{axis}
\end{tikzpicture}
}
\caption{
Histogram distribution of number of relation labels covered by each qualifier label.
}
\label{fig:qualifier_dist}
\end{figure}
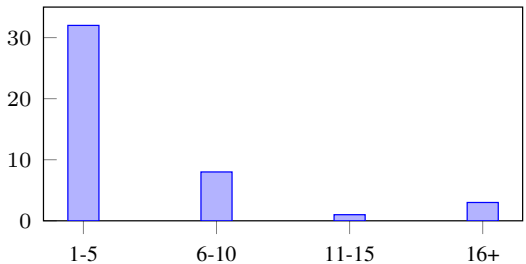

\section{Dataset Details}
\label{sec:data_details}
\paragraph{Dataset Statistics}
Table \ref{tab:data_details} shows the detailed statistics of \dataname{}, such as the number of unique facts and entities, as well as the average number of words in each sentence. 
Table \ref{tab:relations} and Table \ref{tab:qualifiers} show the set of relation and qualifier labels respectively.
For the construction of the dataset, we use the Wikidata which has 594,088 hyper-relational facts and introductions from English Wikipedia which has 4,650,000 articles.

\paragraph{Distant Supervision Example}
In this section, we demonstrate the distant supervision process for fact alignment with a sentence example.
Given the input sentence ``Leonard Parker received his PhD from Harvard University in 1967.'', we first perform entity linking which detects the entity mentions and their Wikipedia IDs: \{(Leonard Parker, Q3271532), (PhD, Q752297), (Harvard University, Q13371)\}.
As the entity linker does not consider dates or numbers, we use the spaCy tool to extract such spans: \{(1967, Date)\}.
Hence, the set of linked entities in the sentence is \{(Leonard Parker, Q3271532), (PhD, Q752297), (Harvard University, Q13371), (1967, Date)\}.
To address the case if the sentence contains unresolved pronouns such as ``he'' or ``she'', we use the Stanford CoreNLP tool to detect and resolve such cases to a suitable entity in the set of linked entities above.
For each hyper-relational fact the in Wikidata knowledge graph, we attempt to align it to the sentence based on the entities in the fact.
If the head entity, tail entity and value entity are all present in the linked entities set of the sentence, then it is a successful alignment.
For example, given the fact (Leonard Parker, Educated At, Harvard University, End Time, 1967) where the head entity, tail entity and value entity is (Leonard Parker, Q3271532), (Harvard University, Q13371) and (1967, Date) respectively, the fact is successfully aligned with the sentence as the three entities are present in the set of linked entities.
If any entities are missing from the set of linked entities, the alignment is unsuccessful and we do not include it in the dataset.
If any sentence does not have any successfully aligned facts, we do not include it in the dataset.

\paragraph{Annotation Challenges}
The human annotation of the dataset may be imperfect due to complexity of the hyper-relational fact structure, diversity of relation and qualifier labels, and possible ambiguous facts.
The hyper-relational facts require annotators to joint consider the relation triplet and qualifier which is more challenging compared to previous datasets which commonly consider the relation between two entities.
On the other hand, the annotators are also required to consider the definitions of a large set of relation and qualifier labels. 
This may pose difficult when some relations or qualifiers are similar in meaning.
Lastly, there may be ambiguous cases where multiple entities are mentioned in relation to a topic and it is not clear which entity is the main subject.

\paragraph{Relation-Specific Qualifiers}
To investigate the link between relation triplets and qualifiers, we plot a histogram distribution in Figure \ref{fig:qualifier_dist}.
A majority (32) of the qualifier labels are each linked to a small number of relation labels (1-5), which suggests that most qualifiers are highly relation-specific.
For example, the ``electoral district'' qualifier label is only linked to the ``candidacy in election'' and ``position held'' relation labels.
On the other hand, a few (3) qualifier labels are each linked to a large number (16+) of relation labels, and not specific to any particular relation.
For example, the ``end time'' qualifier is linked to 35 relation labels.
Hence, it is generally important to consider the interaction between relation triplets and qualifiers in extracting hyper-relational facts.
However, it is not trivial to predict the qualifier only based on the relation, as some qualifier labels are relation-agnostic and it also requires the model to consider the value entity.

\section{Decoding Algorithm}
\label{sec:decode_algo}

\begin{algorithm}
\small
\SetAlgoLined
\PyComment{y\_t: Input entity-relation scores (Eq.3)} \\
\PyComment{y\_q: Input qualifier scores (Eq.5)} \\
\PyCode{} \\
\PyCode{facts = []} \PyComment{Output hyper-relational facts} \\
\PyCode{groups = []} \PyComment{Hyper-relational span groups} \\
\PyCode{} \\
\PyComment{Find and merge adjacent non-null entries} \\
\PyCode{for i,j,k in y\_q.argmax(-1).nonzero():}  \\
\Indp   
    \PyCode{entry = (i,i+1,j,j+1,k,k+1)} \\
    \PyCode{for spans in groups:} \\
    \Indp
        \PyCode{if is\_adjacent(spans, entry):} \\
        \Indp
            \PyCode{merge(spans, entry)} \\
            \PyCode{break} \\
        \Indm
    \Indm
    \PyCode{else:} \\
    \Indp
        \PyCode{groups.append(entry)} \\
    \Indm
\Indm 
\PyCode{} \\
\PyComment{Aggregate relation and qualifier scores} \\
\PyCode{for spans in groups:} \\
\Indp
    \PyCode{i,i2,j,j2,k,k2 = spans} \\
    \PyCode{r\_scores = y\_t[i:i2,j:j2]} \\
    \PyCode{r\_label = r\_scores.mean(0,1).argmax()} \\
    \PyCode{q\_scores = y\_q[i:i2,j:j2,k:k2]} \\
    \PyCode{q\_label = q\_scores.mean(0,1,2).argmax()} \\
    \PyCode{facts.append((spans,r\_label,q\_label))} \\
\Indm
\caption{
Pseudocode of our decoding algorithm in a PyTorch-like style.
}
\label{algo:decoding}
\end{algorithm}

We include the pseudocode algorithm of the proposed decoding method in Algorithm \ref{algo:decoding}.
Note that we can use the nonzero operation to find and merge adjacent non-null entries as it returns the entries sorted in lexicographic order.
This ensures that the order of entries seen in consecutive order if they correspond to the same hyper-relational fact.

\section{Model Costs}
\label{sec:efficiency}
Table \ref{tab:costs} shows a comparison of total training time, inference speed in samples per second and GPU memory usage for the different models. We observe that \modelname{} has a comparable computational cost with the generative and pipeline models. 
This result that our cube-pruning method is effective in ensuring that the model is computationally efficient and practical in real applications.
Note that we compute the statistics for the two-stage pipeline model by summing the time taken and memory used by both stages.

\begin{figure}
\centering
\resizebox{0.75\linewidth}{!}{
\begin{tikzpicture}
\pgfplotsset{width = 6cm, height = 4cm}
    \begin{axis}[
        ymax=70,
        ymin=40,
        ylabel={$F_1$ (\%)},
        xlabel={Fraction of Training Samples Used},
        label style={font=\fontsize{7}{1}\selectfont},
        xtick = {1,2,3,4,5},
        xticklabels = {0.2, 0.4, 0.6, 0.8, 1.0},
        xticklabel style = {font=\fontsize{7}{1}\selectfont},
        yticklabel style = {font=\fontsize{7}{1}\selectfont},
        xtick pos = left,
        ytick pos = left,
        ymajorgrids = true,
        grid style=dashed, 
    ]
    \addplot [mark=square, mark size=1.2pt, color=orange] plot coordinates {
    (1, 47.25) (2, 55.75) (3, 58.01) (4, 62.78) (5, 65.77)};
    \end{axis}
\end{tikzpicture}
\vspace{-2mm}
}
\caption{
The effect of training data size on Dev $F_1$. 
The training set of HyperRED is distantly supervised, while the development and test set are human-annotated.\vspace{-3mm}}
\label{fig:data_size}
\end{figure}
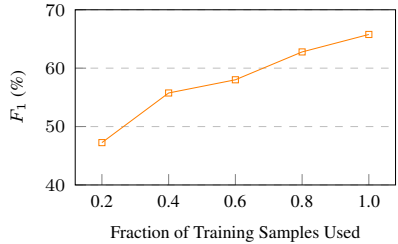

\section{Further Analysis}
\label{sec:further}
\paragraph{Additional Pipeline Results}
For a fair comparison of main results in Section \ref{sec:main_results}, we do not include the pipeline baseline in the large model setting as it would have 680M parameters which is much more than the other models.
On the other hand, we also do not include a BERT-Base version of the pipeline baseline in the main results, as it would have 221M parameters which is not comparable to both the base and large model settings.
Hence, we only include the pipeline baseline using DistilBERT in the main result discussion as it has a comparable parameter count to the base model setting.
However, we include the pipeline baseline with BERT-Base in Table \ref{tab:results2} for reference.

\paragraph{Effect of Pruning}
The main effect of cube-pruning is to reduce the sparsity of the cube entries by retaining the entries which are most likely to be valid entities.
To quantify the effect on sparsity, we measure the cube without pruning to consist of 99.9900\% null entries on average.
When using pruning threshold $m=20$, the cube consists of 99.9098\% null entries on average.
Hence, there is a roughly tenfold increase in the proportion of non-null entries when using pruning.

\paragraph{Effect of Training Data Size}
\label{sec:size}
The \dataname{} training set consists of distantly supervised data which enables large-scale and diverse model training.
However, there may be noisy samples that affect the model performance.
Hence, we aim to study whether the quantity of data can overcome noise in the training set.
As shown in Figure \ref{fig:data_size}, we observe a strictly increasing trend when the size of the training set is increased from 20\% of the original size to 100\% of the original size.
Thus, the results suggest that the quantity of data is still a beneficial factor for model performance despite some noise in the distantly supervised training set.

\begin{table*}[!t]
    \centering
    \resizebox{1\textwidth}{!}{
    \begin{tabular}{lll}
    \toprule
    Wiki ID & Label & Description \\
    \midrule{}
    P6 & head of government & head of the executive power of this town, city, municipality, state, country, or other governmental body \\
    P17 & country & sovereign state of this item (not to be used for human beings) \\
    P19 & place of birth & most specific known (e.g. city instead of country, or hospital instead of city) birth location of a person, animal or fictional cha
    racter \\
    P26 & spouse & the subject has the object as their spouse (husband, wife, partner, etc.). Use "unmarried partner" (P451) for non-married companions \\
    P27 & country of citizenship & the object is a country that recognizes the subject as its citizen \\
    P31 & instance of & that class of which this subject is a particular example and member \\
    P35 & head of state & official with the highest formal authority in a country/state \\
    P39 & position held & subject currently or formerly holds the object position or public office \\
    P40 & child & subject has object as child. Do not use for stepchildren \\
    P47 & shares border with & countries or administrative subdivisions, of equal level, that this item borders, either by land or water. A single common point is enough. \\
    P54 & member of sports team & sports teams or clubs that the subject represents or represented \\
    P69 & educated at & educational institution attended by subject \\
    P81 & connecting line & railway line(s) subject is directly connected to \\
    P97 & noble title & titles held by the person \\
    P102 & member of political party & the political party of which a person is or has been a member or otherwise affiliated \\
    P106 & occupation & occupation of a person; see also "field of work" (Property:P101), "position held" (Property:P39) \\
    P108 & employer & person or organization for which the subject works or worked \\
    P115 & home venue & home stadium or venue of a sports team or applicable performing arts organization \\
    P118 & league & league in which team or player plays or has played in \\
    P127 & owned by & owner of the subject \\
    P131 & located in the administrative & the item is located on the territory of the following administrative entity. \\
    & territorial entity  & \\
    P137 & operator & person, profession, or organization that operates the equipment, facility, or service \\
    P156 & followed by & immediately following item in a series of which the subject is a part \\
    P159 & headquarters location & city, where an organization's headquarters is or has been situated. Use P276 qualifier for specific building \\
    P161 & cast member & actor in the subject production \\
    P166 & award received & award or recognition received by a person, organisation or creative work \\                                                       
P175 & performer & actor, musician, band or other performer associated with this role or musical work \\                                                  
P176 & manufacturer & manufacturer or producer of this product \\                                                                                         P179 & part of the series & series which contains the subject \\                                                                                          
P194 & legislative body & legislative body governing this entity; political institution with elected representatives, such as a parliament/legislature or council \\                                                                                                                                                
P197 & adjacent station & the stations next to this station, sharing the same line(s) \\                                                                  
P241 & military branch & branch to which this military unit, award, office, or person belongs, e.g. Royal Navy \\                             
P276 & location & location of the object, structure or event. In the case of an administrative entity as containing item use P131. \\
P279 & subclass of & next higher class or type; all instances of these items are instances of those items; this item is a class (subset) of that item. \\
P361 & part of & object of which the subject is a part \\
P414 & stock exchange & exchange on which this company is traded \\
P449 & original broadcaster & network(s) or service(s) that originally broadcasted a radio or television program \\
P463 & member of & organization, club or musical group to which the subject belongs. Do not use for membership in ethnic or social groups \\
P466 & occupant & person or organization occupying property \\
    P488 & chairperson & presiding member of an organization, group or body \\                                                                                
P551 & residence & the place where the person is or has been, resident \\                                                                                 
P641 & sport & sport that the subject participates or participated in or is associated with \\
P669 & located on street & street, road, or square, where the item is located. \\                                                                                                          
P710 & participant & person, group of people or organization (object) that actively takes/took part in an event or process (subject). \\                                      
P725 & voice actor & performer of a spoken role in a creative work such as animation, video game, radio drama, or dubbing over \\               
P749 & parent organization & parent organization of an organization, opposite of subsidiaries (P355) \\
P793 & significant event & significant or notable events associated with the subject \\
P800 & notable work & notable scientific, artistic or literary work, or other work of significance among subject's works \\
P1037 & director / manager & person who manages any kind of group \\
P1327 & partner in business or sport & professional collaborator \\
P1346 & winner & winner of a competition or similar event, not to be used for awards \\
P1365 & replaces & person, state or item replaced. Use "structure replaces" (P1398) for structures. \\
P1376 & capital of & country, state, department, canton or other administrative division of which the municipality is the governmental seat \\
P1411 & nominated for & award nomination received by a person, organisation or creative work (inspired from "award received" (Property:P166)) \\
    P1441 & present in work & this (fictional or fictionalized) entity or person appears in that work as part of the narration \\
P1535 & used by & item or concept that makes use of the subject (use sub-properties when appropriate) \\
P1923 & participating team & like 'Participant' (P710) but for teams. For an event like a cycle race or a football match you can use this property to list the teams \\
P3450 & sports season of  & property that shows the competition of which the item is a season. Use P5138 for "season of club or team"
. \\
& league or competition & \\
P3602 & candidacy in election & election where the subject is a candidate \\
P3701 & incarnation of & incarnation of another religious or supernatural being \\
P5800 & narrative role & narrative role of this character (should be used as a qualifier with P674 or restricted to a certain work using P642) \\
P6087 & coach of sports team & sports club or team for which this person is or was on-field manager or coach \\

    \bottomrule
    \end{tabular}
    }
   \caption{List of relation labels in \dataname{}.} 
    \label{tab:relations}
\end{table*}

\begin{table*}[!t]
    \centering
    \resizebox{1\textwidth}{!}{
    \begin{tabular}{lll}
    \toprule
    Wiki ID & Label & Description \\
    \midrule
    P17 & country & sovereign state of this item (not to be used for human beings) \\                                                                         P25 & mother & female parent of the subject. For stepmother, use "stepparent" (P3448) \\                                                                  
P31 & instance of & that class of which this subject is a particular example and member \\                                                                P39 & position held & subject currently or formerly holds the object position or public office \\                                              
P81 & connecting line & railway line(s) subject is directly connected to \\                                                                               
P102 & member of political party & the political party of which a person is or has been a member or otherwise affiliated \\                               
P131 & located in the administrative  & the item is located on the territory of the following administrative entity. \\                                                                                                                                                         
& territorial entity & \\
P155 & follows & immediately prior item in a series of which the subject is a part, preferably use as qualifier of P179 \\
P175 & performer & actor, musician, band or other performer associated with this role or musical work \\
P197 & adjacent station & the stations next to this station, sharing the same line(s) \\
P249 & ticker symbol & identifier for a publicly traded share of a particular stock on a particular stock market or that of a cryptocurrency \\
P276 & location & location of the object, structure or event. In the case of an administrative entity as containing item use P131. \\
P413 & position played on  & position or specialism of a player on a team \\
& team / speciality & \\
P453 & character role & specific role played or filled by subject -- use only as qualifier of "cast member" (P161), "voice actor" (P725) \\
P512 & academic degree & academic degree that the person holds \\
P518 & applies to part & part, aspect, or form of the item to which the claim applies \\                                                                  
P527 & has part & part of this subject; inverse property of "part of" (P361). See also "has parts of the class" (P2670). \\
P577 & publication date & date or point in time when a work was first published or released \\          
P580 & start time & time an event starts, an item begins to exist, or a statement becomes valid \\
P582 & end time & time an item ceases to exist or a statement stops being valid \\                                                             
P585 & point in time & time and date something took place, existed or a statement was true \\                                                             
P642 & of & qualifier stating that a statement applies within the scope of a particular item \\                 
P670 & street number & number in the street address. To be used as a qualifier of Property:P669 "located on street" \\
P708 & diocese & administrative division of the church to which the element belongs \\
P768 & electoral district & electoral district this person is representing, or of the office that is being contested. \\
P805 & statement is subject of & (qualifying) item that describes the relation identified in this statement \\
P812 & academic major & major someone studied at college/university \\
P1114 & quantity & number of instances of this subject \\
P1129 & national team appearances & total number of games officially played by a sportsman for national team \\
P1310 & statement disputed by & entity that disputes a given statement \\
P1346 & winner & winner of a competition or similar event, not to be used for awards \\
P1350 & number of matches  & matches or games a player or a team played during an event. \\
& played/races/starts & \\
P1352 & ranking & subject's numbered position within a competition or group of performers \\                                                              
P1365 & replaces & person, state or item replaced. Use "structure replaces" (P1398) for structures. \\                                                                                                        
P1416 & affiliation & organization that a person or organization is affiliated with (not necessarily member of or employed by) \\                         
P1545 & series ordinal & position of an item in its parent series (most frequently a 1-based index), generally to be used as a qualifier \\                                                          
P1686 & for work & qualifier of award received (P166) to specify the work that an award was given to the creator for \\                                   
P1706 & together with & qualifier to specify the item that this property is shared with \\                                                                
P2453 & nominee & qualifier used with «nominated for» to specify which person or organization was nominated \\                                            
P2868 & subject has role & role/generic identity of the item ("subject"), also in the context of a statement. \\
P3831 & object has role & (qualifier) role or generic identity of the value of a statement ("object") in the context of that statement \\
P3983 & sports league level & the level of the sport league in the sport league system \\                                        
P5051 & towards & qualifier for "adjacent station" (P197) to indicate the terminal station(s) of a transportation line or service in that direction \\
    \bottomrule
    \end{tabular}
    }
   \caption{List of qualifier labels in \dataname{}.} 
    \label{tab:qualifiers}
\end{table*}

\end{document}